\documentclass[conference]{IEEEtran}

\usepackage{tabu}
\usepackage{cite}
\usepackage{graphicx} 
\usepackage[utf8]{inputenc}
\usepackage{algorithm}
\usepackage{algpseudocode}
\usepackage{amsmath}
\usepackage{multirow}
\pdfoutput=1
\input epsf

\begin{document}

\title{GenSample: A Genetic Algorithm for Oversampling in Imbalanced Datasets}

\makeatletter
\newcommand{\linebreakand}{%
  \end{@IEEEauthorhalign}
  \hfill\mbox{}\par
  \mbox{}\hfill\begin{@IEEEauthorhalign}
}
\makeatother



\author{\IEEEauthorblockN{Vishwa Karia}
\IEEEauthorblockA{Center for Smart Health \&\\
Computer Science Department, UCLA\\
Email: vishwakaria2@ucla.edu}
\and
\IEEEauthorblockN{Wenhao Zhang}
\IEEEauthorblockA{Center for Smart Health \&\\
Computer Science Department, UCLA\\
Email: wenhao@ucla.edu}
 \linebreakand
\IEEEauthorblockN{Arash Naeim}
\IEEEauthorblockA{Center for Smart Health \&\\
Department of Medicine, UCLA\\
Email: ANaeim@mednet.ucla.edu
}
\and
\IEEEauthorblockN{Ramin Ramezani}
\IEEEauthorblockA{Center for Smart Health \&\\
Computer Science Department, UCLA\\
Email: raminr@ucla.edu}
}


\maketitle
\thispagestyle{plain}
\pagestyle{plain}
\begin{abstract}
Imbalanced datasets are ubiquitous. Classification performance on imbalanced datasets is generally poor for the minority class as the classifier cannot learn decision boundaries well. However, in sensitive applications like fraud detection, medical diagnosis, and spam identification, it is extremely important to classify the minority instances correctly. In this paper, we present a novel technique based on genetic algorithms, GenSample, for oversampling the minority class in imbalanced datasets. GenSample decides the rate of oversampling a minority example by taking into account the difficulty in learning that example, along with the performance improvement achieved by oversampling it. This technique terminates the oversampling process when the performance of the classifier begins to deteriorate. Consequently, it produces synthetic data only as long as a performance boost is obtained. The algorithm was tested on 9 real-world imbalanced datasets of varying sizes and imbalance ratios. It achieved the highest F-Score on 8 out of 9 datasets, confirming its ability to better handle imbalanced data compared to other existing methodologies. 
\end{abstract}


\IEEEoverridecommandlockouts

\IEEEpeerreviewmaketitle

\section{Introduction}
Building classifiers for imbalanced datasets is a difficult task. A dataset is said to be balanced when it has approximately the same number of samples from all classes. A well-balanced dataset provides a fair view to the classifier, helping it learn decision boundaries without any bias. Generally, the goal of classifiers is to maximize the accuracy of predictions. When a dataset is imbalanced, labeling new samples as belonging to the majority class decreases the likelihood of making mistakes during prediction. For instance, if the minority class makes up just 1\% of the dataset, predicting every data point as belonging to the majority class will lead to a 99\% accuracy. Consequently, the classification of minority samples is highly compromised. However, in sensitive applications like fraud detection, medical diagnosis, detection of defects in production, the minority class with rare instances is of greater interest. Thus, several methods have been proposed to balance datasets before feeding them to the classifier. For the scope of this paper, we restrict our discussion to the binary classification task only. We will assume the minority class to be positive and the majority class to be negative.

The techniques used for balancing datasets can be broadly classified into two categories – Data level and Algorithm level \cite{1}. Data level techniques focus on oversampling the minority class or under-sampling the majority class, either randomly or in a directed manner. Algorithm level techniques aim to alter the costs of the various classes, adjust the decision threshold or use ensemble learning. One of the most popular data level techniques, SMOTE \cite{smote}, creates new minority examples by interpolating between minority observations in the original dataset. An extension of this technique, ADASYN \cite{adasyn}, assigns weights to different minority class examples based on the difficulty in learning those examples. 

In this paper, we present a novel data-level approach based on Genetic Algorithms \cite{ga} that considers both, the difficulty in learning the features of an example and the performance improvement caused by oversampling it, during the process of resampling. The process of oversampling a dataset to make it balanced can be considered to be analogous to a population growing and evolving over time. Since genetic algorithms replicate the process of natural selection and evolution in computation problems, their application to imbalanced datasets could be expected to perform well. This expectation is supported by the findings of this paper which show better classification results in terms of $F_{1}$ Score in 8 of the 9 datasets we experimented with. We believe that it is not necessary to completely balance a dataset for its features to be understood by a classifier. In fact, adding more artificial data than necessary for proper classification can sometimes degrade performance. Thus, our algorithm terminates when it encounters a decline in the $F_{1}$ Score, thereby assuring that the results will not be worse than the original classifier in the case where they cannot be improved.

The rest of this paper is organized as follows: We introduce the Genetic oversampling algorithm, GenSample in Section \ref{sec:algorithm}. We discuss the algorithm in terms of the selection, crossover and mutation operations of Genetic Algorithms. In Section \ref{sec:experiments}, we perform experiments on 9 real-world datasets to evaluate the performance of GenSample. We compare the results with 3 benchmark algorithms, the naive Decision Tree, SMOTE and ADASYN, and show how GenSample shows better performance in general. 

\section{Related Work} \label{sec:relatedwork}
Traditional ways to handle an imbalance in datasets include data-level techniques like under-sampling the majority class or oversampling the minority class, and algorithm-level techniques like cost-sensitive learning and ensemble methods. 

\subsection{Undersampling} Random undersampling eliminates some majority class examples, leading to a better imbalance ratio of the dataset. The concept of Tomek Links, introduced in \cite{tomek}, was one of the first systematic procedures for oversampling, focussing on eliminating the borderline majority examples. Kubat \textit{et al.} extend the concept of Tomek Links to remove majority class samples while leaving the minority class as it is, called `One-sided selection’ \cite{undersample}. Elhassan T. \textit{et al.} also combine random undersampling with Tomek Links to balance datasets \cite{tomek2}. However, there is a tendency of losing important features of the data during undersampling. Thus, it is not popularly used as a standalone technique. Instead, it is either coupled with oversampling or avoided.

\subsection{Oversampling} The simplest oversampling techniques work by duplicating the minority class entities i.e. by generating new data which is a replica of an existing data point. The advantage of this technique is that it is extremely safe. We are only adding the examples which are valid observations, however, by repeatedly presenting them to the classifier, we are helping it learn these rare features better. Chawla \textit{et al.} proposed a novel oversampling technique called SMOTE (Synthetic Minority oversampling Technique) which systematically generates new minority data along the line joining each minority point and one of its nearest neighbors \cite{smote}. Han \textit{et al.} extended the idea of SMOTE to Borderline-SMOTE which focusses on strengthening the data points near the decision boundary, oversampling the borderline points the most \cite{bordersmote}. ADASYN, presented in \cite{adasyn} incorporates the concept of adaptively synthesizing new points depending on the ratio of the majority class samples in the $k$-neighborhood of a point. This idea of distinguishing between different types of examples was also proposed in \cite{4}.

\subsection{Algorithm-Level Approaches} Cost-Sensitive Learning tackles the imbalance problem by penalizing the misclassification of an example based on its class. This allows us to place more emphasis on correctly identifying the instances of the minority class by giving that class a higher weight. Ensemble methods have also proven to be promising for handling imbalanced datasets. For instance, SMOTEBoost oversamples the minority class by leveraging SMOTE at each step of boosting while learning the weak classifier \cite{smoteboost}.

The use of metaheuristic algorithms for handling imbalance in datasets has been rare. Yu \textit{et al.} presented ACOSampling, which uses Ant Colony Optimization for undersampling in DNA microarray data \cite{aco}. A recent algorithm, GASMOTE, proposed by Jiang \textit{et al.} in \cite{gasmote}, augments SMOTE with a genetic algorithm to perform oversampling. In this method, an individual in the population is a sequence of sampling rates for all minority instances. These individuals are evolved until the optimal oversampling rate for each example is reached. 

\section{GenSample Algorithm} \label{sec:algorithm}
As described in [4], minority class examples can be divided into four categories: safe, borderline, rare and outliers. Safe examples are located in homogenous regions of similar examples and are easier to classify. Borderline examples are the ones close to the decision boundary, causing more difficulty during classification. Rare examples are the outlying pairs or triplets of minority class points corresponding to the under-represented minority regions, making them more fit for resampling than the former two types. Outliers, as the name suggests, are the scarce minority examples scattered in regions of majority, representing either noise or a valid, extremely rare subconcept. There are contrasting opinions on how to handle outliers, whether to discard or resample them. However, many studies on medical problems have shown that these outliers are authentic, under-represented minority samples. Hence, GenSample considers these points to be the most difficult to learn, oversampling them the most. 

The main idea of the GenSample algorithm is to iteratively learn which minority samples are best suited for resampling. We set aside one-third of the training data as a validation dataset. The algorithm first generates the \emph{initial population} which consists of all the minority examples from the training data. It then uses the \emph{fitness function} described in section \ref{sec:fitness} to pick the fittest individual of the population as the first parent for \emph{crossover}. The second parent is randomly selected from the k-nearest minority neighbors of the first parent. An interpolation between the two parents like \cite{smote}  produces two children, each of which is evaluated for its fitness. The fitter of the two children replaces the least fit individual of the population. Since each minority example is important to our classifier, we eliminate the least fit individual only from the population, not from the original dataset used for classification. This can be interpreted as considering that individual unfit for reproduction but fit enough to survive. Thus, we can view the population as the set of individuals who participate in crossover. During evaluation, we fit a Decision Tree classifier with the entire training data, not just the population. The fitted model is tested on the validation dataset. Each of the above-mentioned steps is elaborated in the following sections and formally presented in Algorithm \ref{alg:alg}.

\begin{algorithm*}
  \caption{GenSample Algorithm}\label{euclid}
  \label{alg:alg}
  \begin{algorithmic}[1]
    \Procedure{evaluate\_dataset}{$X_{train}$, $X_{test}$, $y_{train}$, $y_{test}$}
    \State Fit a Decision tree Classifier on $X_{train}$
    \State \Return $f\_score(y_{test}, y_{train})$ 
    \EndProcedure
    \Procedure{evaluate\_population\_fitness}{$population$, $X_{train}$, $\Delta Fscore$, $beta$}
      \For{$individual$ in $population$}
        \State Calculate $fitness(individual)$ 
      \EndFor
      \EndProcedure
      \Procedure{GenSample}{$X_{train}$, $beta$}
      \State\textit{Initialize $population$ $\gets$ $\{x | x \in X_{min}\}$}
      \State\textit{Initialize $new\_samples$ $\gets$ 0}
      \State $target\_new\_samples$ = $|X| - 2 \cdot |X_{min}|$ \Comment{Balances the imbalance ratio}
      \State \textsc{evaluate\_population\_fitness}($population$,$X_{train}$, 0, $beta$)
      \While{$new\_samples < target\_new\_samples$ and $curr\_eval\_measure \leq prev\_eval\_measure$}
        \State $explore$ $\gets$ random$([True, False], p = [0.15,0.85])$ 
        \If {$explore$ = $False$}
            \State $fittest_{1}$ $\gets$ $population$.fittest  \Comment{Selection}
        \Else 
            \State $fittest_{1}$ $\gets$ Random Individual in the $population$
        \EndIf
        \State $fittest_{2}$ $\gets$ Random minority neighbor from $k$-nearest neighbors of $fittest1$
        \State Generate $child_{1}$, $child_{2}$
        \Comment{Crossover}
        \State Add $child_{1}$ to dataset
		\State $eval\_measure_{1}$ $\gets$ \textsc{evaluate\_dataset}
		\State Remove $child_{1}$
		\State Add $child_{2}$ to dataset
		\State $eval\_measure_{2}$ $\gets$ \textsc{evaluate\_dataset}
	    \State Remove $child_{2}$
		\State $fitter\_child$ $\gets$ ($eval\_measure_{1} \geq eval\_measure_{2}$) ? $child_{1}$ : $child_{2}$
		\State Replace least fit point in $population$ with $fitter\_child$
		\State Add $fitter\_child$ to dataset
      \EndWhile\label{euclidendwhile}
    \EndProcedure
  \end{algorithmic}
\end{algorithm*}

\subsection{Initial Population}
As mentioned before, the initial population consists of all the minority class samples as individuals. To evaluate the dataset during resampling, we use the $F_{1}$ score as an evaluation metric because it takes into consideration both the precision and recall, giving us a single metric to gauge our performance. 
\subsection{Fitness Function} \label{sec:fitness}
The fitness of an individual depends on the type of minority class example it is, i.e. safe, borderline, rare or outlier, as well as the amount of performance improvement achieved by oversampling it. The more challenging it is to classify an example, the more it should be resampled. Consequently, its fitness value should be high. The fitness function interpolates between the two above measures as follows:
\begin{eqnarray*}
fitness(x) & = & beta \times minority\_label\_weight \\
& & + (1 - beta) \times \Delta Fscore
\end{eqnarray*}

where $beta$ is a constant such that $0 < beta < 1$. $minority\_label\_weight$ depends on the category of the minority class and is calculated using the number of majority class samples in the k-neighborhood of $x$. $\Delta Fscore$ is the change in $F_{1}$ score produced by resampling $x$.

The $minority\_label\_weight$ is assigned as follows:\\
\textbf{If} $0.75 \leq majority\_neighbors\_ratio \leq 1.0$
		    
		    $minority\_label\_weight$ $\gets$ random.uniform(0.8, 1.0)\\
\textbf{Else If} $0.5 \leq majority\_neighbors\_ratio < 0.75$

		    $minority\_label\_weight$ $\gets$ random.uniform(0.6, 0.8)\\
\textbf{Else If} $0.25 \leq majority\_neighbors\_ratio < 1.0$

			$minority\_label\_weight$ $\gets$ random.uniform(0.4, 0.6)\\
\textbf{Else}

		     $minority\_label\_weight$ $\gets$ random.uniform(0.2, 0.4)

Here, $majority\_neighbors\_ratio$ is the ratio of majority samples in the $k$-neighborhood of the point under consideration. The minority label weights used above are based on empirical results. Randomization ensures that samples belonging to the same category do not end up with the same fitness value. 
\subsection{Selection}
After calculating the fitness of all the individuals in the population, the one with the highest fitness is selected as the first parent for crossover. The second parent is randomly picked from the $k$ nearest minority neighbors of the first parent, where $k$ can be selected by cross-validation.
\subsection{Crossover}
The crossover mechanism generates children by interpolating between the parents, like \cite{smote}. If we draw a line joining the two parents, the newly generated sample will lie somewhere on the line segment between the parent points. 
\[
child  =  parent_{1} + (parent_{2} - parent_{1}) \times \lambda \\
\]
where $0 < \lambda < 1$.

However, if the fittest individual is an outlier, its nearest neighbors are going to be very far from it. When we randomly select a point along the line, we might not generate a point close to the outlier at all. To ensure that this does not happen, we produce two children, each one closer to one of the parents:
\begin{eqnarray*}
child_{1}  & = & fittest_{1} + (fittest_{2} - fittest_{1}) \times \lambda \\
child_{2}  & = & fittest_{1} + (fittest_{2} - fittest_{1}) \times (1 - \lambda)
\end{eqnarray*}
Here, $0 < \lambda < 1$. Hence, each child will be closer to one parent than the other. 

The fitness of the children is evaluated by adding them to the dataset one after the other. We then observe which child causes a greater performance improvement, thereby the fitter child replaces the least fit individual in the population. However, the least fit individual is not removed from the dataset because it could contain important information. It is only considered unfit for reproduction.
\subsection{Mutation}
The aim of mutation in a genetic algorithm is to maintain diversity in the population and ensure that the optimization does not get stuck in a local maximum. We use the explore-exploit trade-off of machine learning to prevent premature convergence. The selection function \emph{exploits} its current path by choosing the fittest individual in the population most of the time. But, with a small probability, it might choose to \emph{explore} by picking a random individual from the population as the first parent. This ensures that a possible promising parent with a low fitness value might be given a chance to increase its fitness. It will also ensure that the same individual will not be picked an indefinite number of times as the parent.
\subsection{Termination}
The above-mentioned steps are repeated until one of the following two terminating conditions are met: 
\begin{itemize}
    \item The desired imbalance ratio is reached
    \item Adding a new sample caused a degradation in performance
\end{itemize}
The second condition ensures that we do not add samples beyond what is needed for classification. Oversampling more than necessary can lead to ambiguities in the dataset, making it harder for the classifier to find the decision boundary. Hence, we resample the minority class only as much as needed.

\section{Experiments} \label{sec:experiments}
We evaluated the GenSample algorithm on 9 datasets with different imbalance ratios, sizes as well as the number and types of features. First, the parameter settings used for the experiments are described. Next, we describe the datasets used and the modifications made to them for our binary classification problem. A discussion of the metrics used for evaluation is presented next, and finally, the results of the experiments are put forth.

\subsection{Experimental Setup}
Formally described in Algorithm \ref{alg:alg}, the GenSample algorithm begins by computing the fitness of each individual in the original population. The relative importance of the minority class type and the performance improvement obtained by resampling it are both controlled by the parameter $beta$ of the fitness function. Empirically, the value of $beta$ = 0.75 works the best, though it can also be calculated by cross-validation. We chose not to use cross-validation to reduce the complexity of the algorithm. The value of $k$ for kNN is set to be 5, though $k$=7 also produces good results in many cases.

The GenSample algorithm is compared with the naive Decision Tree (C4.5), SMOTE + Decision Tree and ADASYN + Decision Tree algorithms. We take an average of 100 runs for all the algorithms to obtain stable results. Again, the value of $k$=5 is used for both SMOTE and ADASYN. We oversample the minority class for both the algorithms until the number of minority samples becomes equal to that of the majority samples.

\begin{table*}[t]
\centering
\caption{Summary of Dataset Characteristics}
\begin{tabular}{|c|c|c|c|c|c|}
\hline
Dataset Name & Total Datapoints & Minority Datapoints & Majority Datapoints  & Number of Features & Imbalance Ratio\\ 
\hline
Ionosphere & 351 & 126 & 225 & 34 & 1.8\\
\hline 
Heart & 294 & 106 & 188 & 13 & 1.8\\
\hline
Iris & 150 & 50 & 100 & 4 & 2.0\\
\hline
Parkinson & 195 & 48 & 147 & 22 & 3.1 \\
\hline
Blood Transfusion & 748 & 178 & 570 & 4 & 3.2\\
\hline
Vehicle & 846 & 199 & 647 & 18 & 3.3\\
\hline
CMC & 1473 & 333 & 1140 & 9 & 3.4\\
\hline
Yeast & 1484 & 244 & 1240 & 8 & 5.1\\
\hline
PC1 & 1109 & 77 & 1032 & 21 & 13.4 \\
\hline
\end{tabular}
\label{tab:1}
\end{table*}

\subsection{Datasets}
We tested the algorithm on 9 datasets commonly used in the literature for benchmarking. For each experiment, we randomly divide the data into 50\% training and 50\% testing datasets. Their attributes are summarized in Table \ref{tab:1}. Most of these datasets are publicly available on the UCI Machine Learning Repository \cite{uci}. We made a few modifications to these datasets for the binary classification problem similar to other literary experimental setups for such problems. They are described below:

\subsubsection{Ionosphere Dataset} 
This dataset \cite{uci} consists of radar data collected by a system in Goose Bay, Labrador with 2 classes and 34 features. There are 225 `good radar' instances and 126 `bad radar' instances. Thus, we choose `good radar' as the majority class and `bad radar' as the minority class. 

\subsubsection{Heart Dataset}
The Heart Dataset \cite{uci} is a binary dataset that predicts the presence of heart diseases in patients using 13 attributes like age, sex and blood sugar. The presence of heart disease is rare and constitutes about one-third of the data points. 

\subsubsection{Iris Dataset}
This is a 3 class dataset \cite{uci} which uses 4 features to classify an iris plant into one of the categories from `Iris-versicolor', `Iris-setosa and `Iris-virginica'. Each of these classes have 50 samples each. We choose `Iris-virginica' as the minority class and collapse the other two into the majority class. Thus, we get a skewed dataset with 50 minority and 100 majority samples.

\subsubsection{Parkinson Dataset}
This dataset \cite{parkinson} uses 22 attributes to differentiate people with Parkinson's Disease (PD) from those without. There are 48 positive examples of people diagnosed with PD, hence that is chosen as the minority class. The majority class has 147 examples, resulting in an imbalance ratio of 3.1.

\subsubsection{Blood Transfusion Dataset}
This dataset \cite{blood dataset} presents blood donation statistics where each data point represents an individual. The data points are divided into two classes based on whether the individual donated blood in March 2007 using 4 features. We select `yes' as the minority class with 178 data points and `no' as the majority class with 570 data points.

\subsubsection{Vehicle Dataset}
This dataset \cite{uci} uses 2D silhouettes of objects in the form of an image to classify the kind of 3D object it is: a double-decker bus, Chevrolet van, Saab 9000 and an Opel Manta 400. We collapse the bus, Saab, and Opel into the negative class and use the van as the positive class. This results in 199 positive examples and 647 negative examples, giving an imbalance ratio of 3.3.

\subsubsection{CMC Dataset}
This dataset \cite{uci} tries to predict the current contraceptive method choice of women from the following 3 categories: 1: No-use, 2: Long-term, 3: Short-term. It has a total of 1473 examples with 9 features. `Long-term' is selected as the minority class and has 333 samples. The other 2 classes are combined into the majority class with 1140 samples.
    
\subsubsection{Yeast Dataset}
The Yeast Dataset \cite{uci} classifies the localization site of protein into one of `MIT', `CYT', `NUC', `ME3', `ME2', `ME1', `EXC', `VAC', `POX', `ERL' classes. We choose `MIT' as the minority class with 244 data points and the rest are combined into the majority class with 1240 data points. The classification is done using 8 features.

\subsubsection{PC1 Dataset}
It is one of the NASA Metrics Data Program defect data sets \cite{pc1 dataset}. It is highly skewed with 1032 majority points and only 77 minority points, leading to an imbalance ratio of 13.4. Each example is represented by 21 features.

\begin{table*}
\centering
\caption{Experimental Results and Performance Comparisons \protect\footnote{Winning times do not add up to 8 due to the presence of ties.} }
\begin{tabular}{|c|c|c|c|c|c|c|c|}

\hline
Dataset & Algorithm & Precision & Recall & $F_{1}$ Score & AUC & Accuracy & Geometric Mean\\ 
\hline
\multirow{4}{*}{Ionosphere} & Decision Tree & 0.82 & 0.80 & 0.81 & 0.85 & 0.87 & 0.85 \\\cline{2-8}
                                & SMOTE & 0.80 & 0.80 & 0.80 & 0.85 & 0.86 & 0.85 \\\cline{2-8}
                                & ADASYN & 0.79 & 0.81 & 0.80 & 0.85 & 0.86 & 0.84 \\\cline{2-8}
                                & GenSample & \textbf{0.82} & \textbf{0.82} & \textbf{0.82} & \textbf{0.86} & \textbf{0.87} & \textbf{0.86} \\\cline{2-8}
\hline
\multirow{4}{*}{Heart} & Decision Tree & 0.64 & 0.66 & 0.65 & 0.72 & 0.74 & 0.72\\\cline{2-8}
                                & SMOTE & 0.64 & 0.65 & 0.64 & 0.72 & 0.74 & 0.72 \\\cline{2-8}
                                & ADASYN & 0.65 & 0.66 & 0.65 & 0.72 & 0.74 & 0.72\\\cline{2-8}
                                & GenSample & \textbf{0.66} & \textbf{0.66} & \textbf{0.66} & \textbf{0.73} & \textbf{0.75} & \textbf{0.73} \\\cline{2-8}
\hline
\multirow{4}{*}{Iris} & Decision Tree & 0.92 & 0.91 & 0.91 & 0.93 & 0.94 & 0.93 \\\cline{2-8}
                                & SMOTE & 0.92 & 0.91 & 0.91 & 0.94 & 0.94 & 0.94\\\cline{2-8}
                                & ADASYN & 0.92 & 0.91 & 0.91 & 0.94 & 0.94 & 0.93 \\\cline{2-8}
                                & GenSample & \textbf{0.93} & \textbf{0.93} & \textbf{0.92} & \textbf{0.94} & \textbf{0.95} & \textbf{0.94} \\\cline{2-8}
\hline
\multirow{4}{*}{Parkinson} & Decision Tree & 0.65 & 0.65 & 0.64 & 0.77 & 0.83 & 0.76 \\\cline{2-8}
                                & SMOTE & 0.64 & 0.69 & 0.66 & 0.78 & 0.82 & 0.77\\\cline{2-8}
                                & ADASYN & 0.62 & 0.68 & 0.64 & 0.77 & 0.82 & 0.76 \\\cline{2-8}
                                & GenSample & \textbf{0.66} & \textbf{0.69} & \textbf{0.66} & \textbf{0.78} & \textbf{0.83} & \textbf{0.77} \\\cline{2-8}
\hline
\multirow{4}{*}{Blood Transfusion} & Decision Tree & 0.38 & 0.32 & 0.34 & 0.58 & 0.71 & 0.52 \\\cline{2-8}
                                & SMOTE & 0.35 & 0.42 & 0.38 & 0.59 & 0.56 & 0.56 \\\cline{2-8}
                                & ADASYN & 0.34 & 0.44 & 0.38 & 0.59 & 0.66 & 0.57 \\\cline{2-8}
                                & GenSample & \textbf{0.38} & 0.32 & 0.34 & 0.58 & \textbf{0.72} & 0.52 \\\cline{2-8}
\hline
\multirow{4}{*}{Vehicle} & Decision Tree & 0.83 & 0.84 & 0.83 & 0.89 & 0.92 & 0.89 \\\cline{2-8}
                                & SMOTE & 0.82 & 0.85 & 0.84 & 0.90 & 0.92 & 0.89 \\\cline{2-8}
                                & ADASYN & 0.82 & 0.86 & 0.84 & 0.90 & 0.92 & 0.90 \\\cline{2-8}
                                & GenSample & \textbf{0.84} & 0.84 & \textbf{0.84} & \textbf{0.90} & \textbf{0.93} & \textbf{0.90} \\\cline{2-8}
\hline
\multirow{4}{*}{CMC} & Decision Tree & 0.82 & 0.78 & 0.80 & 0.59 & 0.70 & 0.55 \\\cline{2-8}
                                & SMOTE & 0.82 & 0.73 & 0.77 & 0.59 & 0.67 & 0.57\\\cline{2-8}
                                & ADASYN & 0.81 & 0.72 & 0.76 & 0.58 & 0.66 & 0.56 \\\cline{2-8}
                                & GenSample & 0.81 & \textbf{0.79} & \textbf{0.80} & \textbf{0.59} & \textbf{0.70} & 0.55 \\\cline{2-8}
\hline
\multirow{4}{*}{Yeast} & Decision Tree & 0.46 & 0.50 & 0.48 & 0.69 & 0.82 & 0.66\\\cline{2-8}
                                & SMOTE & 0.43 & 0.53 & 0.47 & 0.70 & 0.80 & 0.68\\\cline{2-8}
                                & ADASYN & 0.41 & 0.54 & 0.46 & 0.69 & 0.79 & 0.67\\\cline{2-8}
                                & GenSample & \textbf{0.47} & 0.50 & \textbf{0.48} & 0.69 & \textbf{0.82} & 0.66 \\\cline{2-8}
\hline
\multirow{4}{*}{PC1} & Decision Tree & 0.32 & 0.34 & 0.33 & 0.64 & 0.90 & 0.56 \\\cline{2-8}
                                & SMOTE & 0.27 & 0.40 & 0.32 & 0.66 & 0.88 & 0.61 \\\cline{2-8}
                                & ADASYN & 0.26 & 0.40 & 0.32 & 0.66 & 0.88 & 0.60\\\cline{2-8}
                                & GenSample & \textbf{0.33} & 0.35 & \textbf{0.34} & 0.65 & \textbf{0.91} & 0.57 \\\cline{2-8}
\hline
\multirow{4}{*}{Winning Times} & Decision Tree & 3 & 1 & 2 & 1 & 4 & 0\\\cline{2-8}
                                & SMOTE & 1 & 2 & 3 & 7 & 0 & 5 \\\cline{2-8}
                                & ADASYN & 0 & 5 & 2 & 4 & 0 & 2 \\\cline{2-8}
                                & GenSample & \textbf{8} & \textbf{5} & \textbf{8} & 6 & \textbf{9} & 5\\\cline{2-8}
\hline
\end{tabular}
\label{tab:results}

\end{table*}

\subsection{Evaluation Metrics}
Overall Accuracy (OA) is one of the most common metrics for classification tasks in Machine Learning. It is defined as the ratio of the number of correct predictions to the total number of predictions. 
\[
\text{Accuracy} = \frac{\text{Number of correct predictions}}{\text{Total number of predictions}}
\]
In terms of positives and negatives, we can rewrite the definition as:
\[
\text{Accuracy} = \frac{TP + TN}{TP + TN + FP + FN}
\]
However, when the dataset is imbalanced, accuracy is not an effective measure of a classifier's performance. Suppose we have 100 data points; 95 of them belong to one class, and the rest 5 to another. The classifier, having seen so many examples of the majority class, tends to predict all the samples as belonging to the majority class. It will achieve a 95\% accuracy in this case but will have a terrible performance for the minority class. Since the rare examples are of greater interest in most applications, the high accuracy rate will not be indicative of the class-wise performance of the classifier. This phenomenon, also known as the `Accuracy Paradox', motivates the use of the following additional metrics to evaluate classifiers. Nonetheless, we report the overall accuracy of the classifiers to examine the effect of oversampling on it.
\subsubsection{Precision}
Precision is defined as the number of samples that are actually positive out of the ones identified as positive by the classifier. Precision helps us examine how accurate the claims of our classifier on the positive class are.
\[
\text{Precision} = \frac{TP}{TP + FP}
\]
\subsubsection{Recall}
Recall can be defined as the number of samples correctly identified as positive among the true positive ones. It is a very important metric in sensitive applications where it would be risky to not identify the rare instances. To illustrate, if the minority class corresponds to the presence of tumors in patients, we would not want to risk a patient's life by not diagnosing their tumor when it exists.
\[
\text{Recall} = \frac{TP}{TP + FN}
\]
\subsubsection{$F_{1}$ Score}
$F_{1}$ Score is the harmonic mean of Precision and Recall. Since the use of a  harmonic mean instead of an average punishes extremities in the precision and recall values, $F_{1}$ Score is an excellent metric for creating a balanced classification model.
\[
\text{$F_{1}$ Score} = \frac{2 \cdot Precision \cdot Recall}{Precision + Recall}
\]
\subsubsection{Geometric Mean}
The geometric Mean of the accuracies of the positive and negative classes is more effective at evaluating classifiers which are dealing with imbalanced data. It can be calculated as follows:
\[
\text{G mean} = \sqrt{\text{Positive Accuracy} \times \text{Negative Accuracy}}
\]
 
The Area Under ROC Curve (or AUC) is another metric for evaluating classifiers. However, as demonstrated in \cite{auc}, when the dataset is imbalanced, the AUC does not do a very good job of capturing the relative performance of two models. Hence, we chose not to report it.

\subsection{Experimental Results and Discussion}
Table \ref{tab:results} presents the results of evaluating GenSample and the other benchmark algorithms- naive decision tree, SMOTE + decision tree and ADASYN + decision tree, on the 9 datasets mentioned previously. The best performance metric for each algorithm is highlighted in the table. In the end, the number of times each algorithm wins in performance over all the datasets is tabulated for ease of comparison similar to \cite{adasyn}.

The first conclusion we can draw from the table is that the overall accuracy is always better with GenSample. This is an important result because it shows that our algorithm tries not to compromise on the accuracy of the majority class to improve that of the minority. The $F_{1}$ Score shows an improvement in all the datasets except for Blood Transfusion. Since GenSample terminates when the performance starts decreasing, we can also observe that even when the $F_{1}$ Score is not the best of the 4 algorithms, it is not less than the naive Decision tree. Thus, our algorithm will at least ensure that performance does not degrade when it cannot be improved. This is not guaranteed by SMOTE and ADASYN. This result can be better visualized in Figure \ref{fig:fscore}. 

\begin{figure}
\centering
  \includegraphics[width=0.8\linewidth]{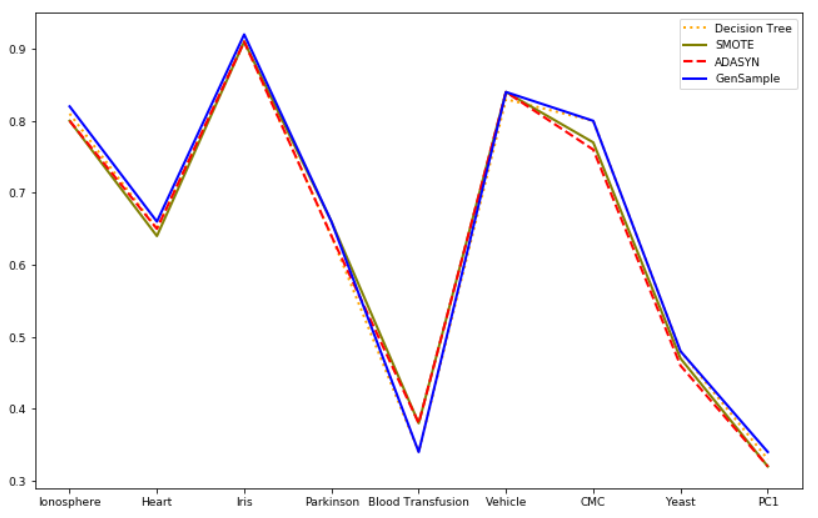}
  \caption{Comparison of $F_{1}$ Score on all datasets}
  \label{fig:fscore}
\end{figure}

The winning times in Table \ref{tab:results} also show that GenSample is generally high on precision. It has better performance in terms of precision in 8 out of the 9 datasets with a minimal number of ties. Even when the precision value is not the highest, it is only slightly less than the best performer. GenSample also achieves good results in terms of recall and geometric mean, doing better half of the time. Similar to $F_{1}$ Score, the performance is never worse than the decision tree.

\section{Conclusion and Future Work}
This paper presents GenSample, a genetic algorithm for handling imbalance in datasets. This algorithm generates synthetic minority data points based on the difficulty in learning a sample point and the performance improvement achieved by oversampling it. The algorithm terminates when the desired imbalanced ratio is reached or a performance deterioration was caused by adding a synthetic data point to the dataset. Due to the early termination condition, the algorithm always ensures that the classification performance does not degrade when it cannot be further improved. We investigate the behavior of GenSample by evaluating it on 9 commonly used imbalanced datasets with 6 different metrics. We observe that for 8 of the 9 datasets, the $F_{1}$ Score and precision are better for Gensample, and the overall accuracy of GenSample is always the highest. Moreover, the recall and geometric mean have the highest value more than 50\% of the time. 

In the future, we will be examining other heuristics which can lead to better results. A promising avenue of research is to investigate the effectiveness of GenSample by combining it with ensemble methods. The data-level techniques have shown considerable improvement when augmented with boosting, so a similar enhancement can be expected from the combination of GenSample and boosting. 

According to the `No Free Lunch Theorem' \cite{nfl}, no single model can work the best for every problem. Thus, although the results for GenSample are not the best for every dataset and metric, they definitely indicate that GenSample holds promising results in the field of imbalanced learning.

\end{document}